\documentclass[lettersize,journal]{IEEEtran}
\usepackage{cite}
\usepackage{amsmath,amsfonts}
\usepackage{algorithm}
\usepackage{array}
\usepackage{textcomp}
\usepackage{stfloats}
\usepackage{url}
\usepackage{verbatim}
\usepackage{graphicx}
\usepackage{algpseudocode}
\usepackage{booktabs}
\usepackage{multirow}
\usepackage[switch,pagewise]{lineno}
\usepackage{orcidlink}
\usepackage[utf8]{inputenc}
\usepackage[caption=false]{subfig}
\usepackage{soul,xcolor}
\usepackage{tablefootnote}
\usepackage{threeparttable} 

\definecolor{airforceblue}{rgb}{0.36, 0.54, 0.66}
\definecolor{MyDarkRed}{rgb}{0.8,0.02,0.02}
\definecolor{MyPurple}{RGB}{111,0,255}
\definecolor{MyRed}{rgb}{1.0,0.0,0.0}
\definecolor{MyGold}{rgb}{0.75,0.6,0.12}
\definecolor{MyDarkgray}{rgb}{0.66, 0.66, 0.66}
\definecolor{MyPink}{rgb}{0.9, 0.33, 0.5}
\definecolor{MyCyan}{rgb}{0., 0.4, 0.4}
\definecolor{AbsoluteColor}{rgb}{0.76, 0.2, 0.2}
\definecolor{DeltaColor}{rgb}{0.87, 0.72, 0.3}
\definecolor{RelativeColor}{rgb}{0.04, 0.33, 0.58}
\definecolor{StanfordRed}{rgb}{0.549, 0.082, 0.082}
\definecolor{BITBrown}{HTML}{A13E0B}
\definecolor{BITGreen}{HTML}{006C39}
\definecolor{MyBlue}{rgb}{0.082, 0.439, 0.910}
\definecolor{MyGreen}{rgb}{0.204, 0.659, 0.325}
\definecolor{MyOrange}{rgb}{0.969, 0.584, 0.294}

\let\titleold\title
\renewcommand{\title}[1]{\titleold{#1}\newcommand{\thetitle}{#1}}
\def\maketitlesupplementary
   {
   \newpage
       \twocolumn[
        \centering
        \Large
        \vspace{0.5em}Supplementary Material \\
        \vspace{1.0em}
       ] 
   }

\begin{document}

\title{ActiveSplat: High-Fidelity Scene Reconstruction\\ through Active Gaussian Splatting}
\markboth{IEEE Robotics and Automation Letters. Preprint Version. Accepted May, 2025}
{Li \MakeLowercase{\textit{et al.}}: ActiveSplat} 

\author{
Yuetao Li$^{1,2*}$, Zijia Kuang$^{2*}$, Ting Li$^{2}$, Qun Hao$^{1}$, Zike Yan$^{2\dagger}$, Guyue Zhou$^{2,3}$, Shaohui Zhang$^{1\dagger}$%
\thanks{
Manuscript received February 22, 2025; accepted May 27, 2025. This paper was recommended for publication by Editors Tamim Asfour and Sven Behnke upon evaluation by the Associate Editor and reviewers.\\
This work was supported by Development and Ecosystem Construction of Robot Simulation and End-to-End Framework, National Key Research and Development Program of China (2021YFC2202404), and National Natural Science Foundation of China (62275020).\\
$^{*}$Equal contributions, $^{\dagger}$Equal advising.\\
$^{1}$School of Optics and Photonics, Beijing Institute of Technology, Beijing 100081, China.\\
$^{2}$Institute for AI Industry Research (AIR), Tsinghua University, Beijing 100084, China.\\
$^{3}$School of Vehicle and Mobility, Tsinghua University, Beijing 100084, China.\\
Digital Object Identifier (DOI): 10.1109/LRA.2025.3580331.
}
}

\maketitle

\begin{abstract}
We propose ActiveSplat, an autonomous high-fidelity reconstruction system leveraging Gaussian splatting. Taking advantage of efficient and realistic rendering, the system establishes a unified framework for online mapping, viewpoint selection, and path planning. The key to ActiveSplat is a hybrid map representation that integrates both dense information about the environment and a sparse abstraction of the workspace. Therefore, the system leverages sparse topology for efficient viewpoint sampling and path planning, while exploiting view-dependent dense prediction for viewpoint selection, facilitating efficient decision-making with promising accuracy and completeness. A hierarchical planning strategy based on the topological map is adopted to mitigate repetitive trajectories and improve local granularity given limited time budgets, ensuring high-fidelity reconstruction with photorealistic view synthesis. Extensive experiments and ablation studies validate the efficacy of the proposed method in terms of reconstruction accuracy, data coverage, and exploration efficiency. 
The released code will be available on our project page\footnote{\href{https://li-yuetao.github.io/ActiveSplat/}{https://li-yuetao.github.io/ActiveSplat/}}.
\end{abstract}


\section{Introduction}
\label{sec:introduction}
\IEEEPARstart{F}{ine-grained} reconstruction of three-dimensional environments has long been a central research focus in robotics, computer vision, and computer graphics. Within the robotics community, there is a growing demand for high-fidelity digitization of the physical world, not only to facilitate immersive applications like teleoperation~\cite{patil2024radiance}, but also to narrow the sim-to-real gap, advancing generalizable robot autonomy through photorealistic simulation~\cite{torne2024reconciling}.

Recent progress in differentiable rendering has significantly improved the quality of the reconstructed environments. Neural radiance fields (NeRF)~\cite{mildenhall2021nerf} and its variant~\cite{barron2023zip} leverage neural networks as compact scene representations, using volume rendering to synthesize high-quality novel views. However, the computational inefficiencies caused by volume integration along rays pose challenges in terms of memory and processing. To address these limitations, Gaussian splatting~\cite{kerbl20233d,Huang2DGS2024} has been introduced, enabling efficient rasterization and achieving promising rendering quality through $\alpha$-blending. Despite these advances, scene-specific and data-dependent optimization makes these methods highly sensitive to noise and artifacts, which often emerge due to insufficient view coverage, especially without ground-truth supervision during data collection.

\begin{figure}
  \centering
  \includegraphics[width=\linewidth]{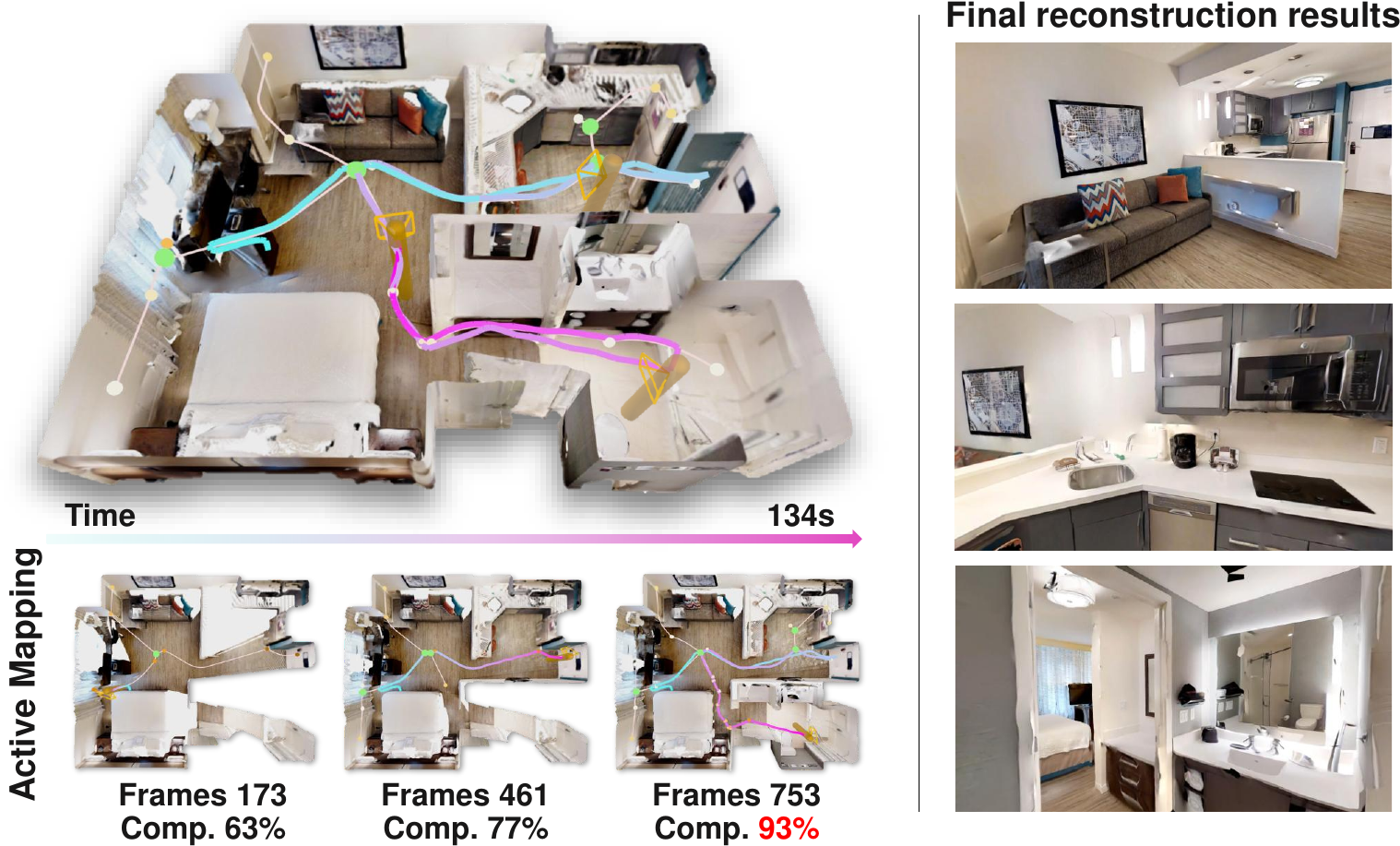}
  \caption{The agent explores the environment autonomously to build a 3D map on the fly. The integration of a Gaussian map and a Voronoi graph ensures efficient and complete exploration, resulting in high-fidelity reconstruction and a high completion ratio (Comp. \%).}
  \label{fig:figure1}
\end{figure}

In this work, we aim to address these issues through active mapping, where a mobile agent reconstructs the environment on the fly, assesses the instant quality of the map, and plans its path to cover the entire environment. We find Gaussian splatting to be particularly suitable for high-fidelity active mapping, owing to its capability for \emph{view-dependent dense predictions}. This characteristic enables the system to efficiently and accurately extract the workspace\footnote{The workspace refers to the navigable free space from a top-down perspective (see Sec.~\ref{subsec:hybrid_map_updating}).} within the horizontal plane, while also quantifying data coverage in a unified manner by splatting Gaussians of interest onto the actively sampled views. The proposed system, dubbed~\textbf{ActiveSplat}, incrementally updates a renderable Gaussian map through gradient-based optimization, progressively refining and completing the scene representation with high fidelity.

To balance reconstruction accuracy and exploration efficiency, our system adopts a hybrid map representation inspired by~\cite{Kuang2024iros}, but replaces the volumetric neural fields with explicit 3D Gaussians, enabling significantly faster convergence and real-time rendering essential for online mapping. A set of 3D Gaussians is maintained as a dense map to provide view-dependent dense predictions, while a Voronoi graph is extracted as a topological map to represent the abstraction of the workspace. Sparse yet representative view positions are derived from this graph, guiding the agent to extend the boundaries of the workspace. Meanwhile, the viewing orientation at each position is determined by view-dependent completeness measures of the Gaussian map. Based on this, a viewpoint decoupling method is proposed to reduce the infinite number of possible viewpoints in free space to a manageable set of positions and rotation angles, ensuring efficient and safe traversal. Additionally, a hierarchical planning strategy based on the topological map is employed to reduce redundant trajectories during global exploration and improve the overall efficiency of the autonomy process. The key contributions of the letter can then be categorized as follows:

\begin{itemize}
    \item A novel system that actively splats Gaussians of interest to build a unified, autonomous, and high-fidelity reconstruction system.
    \item A hybrid map representation combining dense predictions with Gaussians and sparse abstraction of Voronoi graph for comprehensive viewpoint selection and safe path planning.
    \item A hierarchical planning strategy based on the Voronoi graph, which prioritizes local areas to minimize redundant exploration, decoupling viewpoint selection to balance exploration efficiency and reconstruction accuracy.
\end{itemize}

\section{Related Work}
\label{sec:related_works}

\subsection{Autonomous Exploration}
\label{sec:autonomous_exploration}
In the robotic community, autonomous exploration aims to best acquire observations to cover the entire space traversed by the robot. Existing strategies seek to balance exploration completeness and efficiency, and can be broadly categorized as frontier-based methods and sampling-based methods. Frontier-based methods~\cite{yamauchi1997frontier,umari2017autonomous} focus on expanding the exploration area by navigating to the boundary between explored and unexplored regions until full coverage is achieved. However, these methods rely on the discrete grid representation to discern the decision boundary, thus lacking adaptive granularity given diverse geometry complexity. In contrast, sampling-based methods~\cite{schmid2020efficient, Yan2023iccv, Kuang2024iros} sample candidate viewpoints and prioritize those that maximize uncertainty reduction or expected information gain, thus improving scene coverage by reducing environmental uncertainty. Efforts are made to design proper sampling strategies for efficiency and precise scoring techniques given the samples. TARE~\cite{cao2021tare} introduces a hierarchical strategy for LiDAR-based exploration, where a local subspace is traversed at a fine-grained level while global target goals at a coarse level are maintained, thus achieving a balance between exploration efficiency and mapping completeness. Similarly,~\cite{selin2019efficient} adopts fine-grained next-best-view planning for local exploration, while leveraging frontier-based strategies for global coverage. Topology maps have become widely used as sparse scene abstractions in autonomous exploration~\cite{Kuang2024iros, dong2024fast}. Recently, most relevant to ours is~\cite{Kuang2024iros}, which also leverages a hybrid representation containing a dense neural map and a topology map for exploration. However, the neural map suffers from slow convergence and inefficient volume rendering. We, on the other hand, leverage the efficient optimization and rendering of Gaussian primitives to achieve reconstruction with high fidelity. Furthermore, we adopt a decoupled viewpoint selection to separately address translation and rotation during exploration, along with a topology-based hierarchical planning strategy, ultimately achieving a balance between exploration efficiency and completeness.

\subsection{High-Fidelity Scene Reconstruction}
\label{sec:high_fidelity_scene_reconstruction}
Recent progress in differentiable rendering attracts significant attention in the research community. Parameterized by implicit NeRF representations~\cite{mildenhall2021nerf,barron2023zip} or explicit 2D/3D Gaussian representations~\cite{kerbl20233d,Huang2DGS2024}, photorealistic images of novel views can be rendered with promising efficiency. Gradient-based optimization has also been applied in an online setting to incrementally update the neural map~\cite{Zhu2023NICER,jiang2023h} or Gaussian parameters~\cite{keetha2024splatam,matsuki2024gaussian,wei2024gsfusion} through differentiable rendering. Recently, continual learning of the neural map has turned into an active fashion through uncertainty-guided autonomous exploration~\cite{Yan2023iccv,Kuang2024iros, Georgakis2022icra, feng2024naruto, pan2022activenerf}. However, the neural representation has to sacrifice model capacity to achieve real-time performance, requiring inevitable tradeoffs between accuracy and convergence efficiency with different network architectures. In contrast, we adopt a set of Gaussian primitives as the scene representation, which allows consistent optimization of Gaussian parameters in an online update or offline post-processing setting. Relevant works have also developed Gaussian splatting-based mapping systems using unmanned ground vehicle (UGV)~\cite{liu2024beyond, jin2024gs, xu2024hgs, tao2024rt} and unmanned aerial vehicle (UAV)~\cite{chen2025splat, jin2025activegs} platforms. Notably, both the safe navigation system~\cite{liu2024beyond} and HGS-Planner~\cite{xu2024hgs} build upon FisherRF~\cite{jiang2024fisherrf}, which utilizes Fisher information for quantifying the uncertainty of Gaussians. ActiveGS~\cite{jin2025activegs} tackles similar challenges by modeling Gaussian confidence and integrating splatting with voxel maps to improve scene reconstruction. The major difference lies in our proposed hybrid map representation, which enforces safe and hierarchical path planning based on a topological map.

\begin{figure*}
  \centering
  \includegraphics[width=0.8\textwidth]{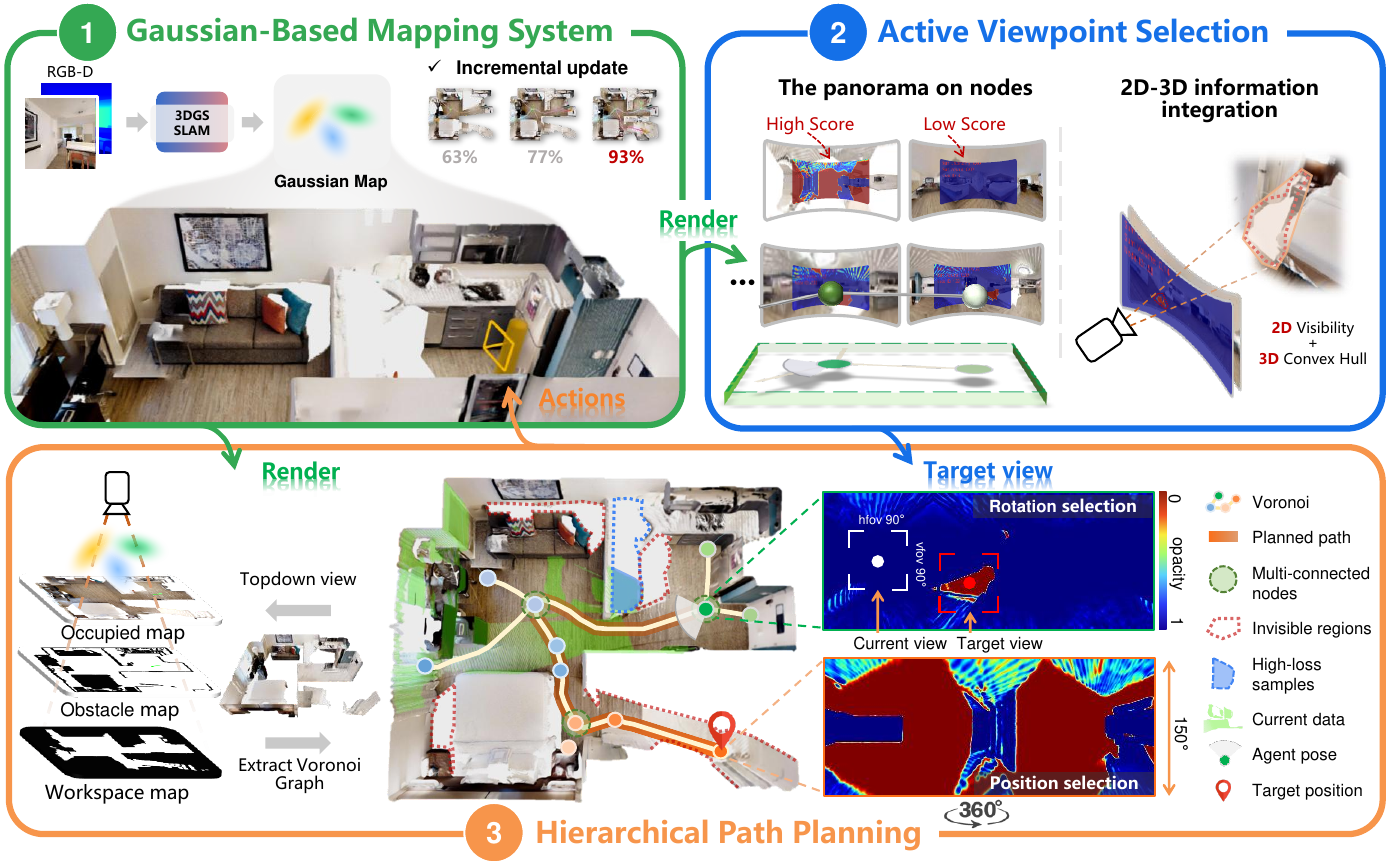}
  \vspace{-2mm}
  \caption{\textbf{Overview of ActiveSplat}: The proposed active mapping system achieves high-fidelity reconstruction through a perception-action closed loop, utilizing a hybrid map that combines dense Gaussians with topological abstractions. Splatting Gaussians of interest on the fly provides a consistent approach for \textcolor{MyGreen}{online map updating}, \textcolor{MyBlue}{viewpoint selection}, and \textcolor{MyOrange}{path planning}. \textbf{Note:} Subregions are distinguished by node color, with node scores indicated by color intensity.}
  \label{fig: pipeline}
  \vspace{-0.2cm}
\end{figure*}
\section{Methodology}
\label{sec:methodology}

Our ActiveSplat system is a Gaussian splatting-based active mapping framework that aims to maximize scene completeness and reconstruction accuracy through autonomous exploration using posed RGB-D input. The hybrid map and hierarchical planning are introduced to improve exploration efficiency within a limited number of steps, balancing the tradeoff between exploration path length and reconstruction completeness. The overview of the system is illustrated in Fig.~\ref{fig: pipeline}, which shows that Gaussians of interest are splatted onto the image plane, serving as a consistent technique for online map updating (Sec.~\ref{subsec:hybrid_map_updating}), viewpoint selection (Sec.~\ref{subsec:active_viewpoint_selection}), and path planning (Sec.~\ref{subsec:hierarchical_plan}).

\subsection{Hybrid Map Updating}
\label{subsec:hybrid_map_updating}
Central to the proposed ActiveSplat system is a hybrid map representation containing both Gaussian primitives, which allow dense prediction, and a topological structure, which provides sparse abstraction of the workspace. A Gaussian primitive is an explicit representation parameterized by color $\mathbf{c}$, center position $\boldsymbol{\mu}$, anisotropic covariance $\Sigma$, and opacity $o$, where the influence of each Gaussian can be expressed as:
\begin{equation}
\label{eq:opacity}
    f_i(\mathbf{u}_k) = o \cdot \exp\left( -\frac{1}{2} (\mathbf{x}(\mathbf{u}_k) - \boldsymbol{\mu}_i)^\top \Sigma^{-1} (\mathbf{x}(\mathbf{u}_k) - \boldsymbol{\mu}_i) \right).
\end{equation}

View synthesis can then be implemented through splatting given the Gaussian map and a camera pose, where the color of each pixel $\mathbf{u}$ is linearly affected by the projected 3D Gaussians as:
\begin{equation}
\label{eq:render_rgb}
\hat{C}_k=\sum_{i=1}^{n_k}\mathbf{c}_if_i(\mathbf{u}_k)\prod_{j=1}^{i-1}\left(1-f_j(\mathbf{u}_k)\right).
\end{equation}

Similarly, the differentiable rendering can also be applied for depth and visibility (accumulated opacity) estimation:
\begin{equation}
\label{eq:render_depth}
    \hat{D}_k=\sum_{i=1}^{n_k} d_if_i(\mathbf{u}_k)\prod_{j=1}^{i-1}\left(1-f_j(\mathbf{u}_k)\right),
\end{equation}
\begin{equation}
\label{eq:render_opacity}
\hat{O}_k=\sum_{i=1}^{n_k}f_i(\mathbf{u}_k)\prod_{j=1}^{i-1}\left(1-f_j(\mathbf{u}_k)\right),
\end{equation}
where $d_i$ is the depth of the Gaussian center in the camera coordinate.

The optimization of the Gaussian map is performed given photometric and geometric losses defined in~\cite{keetha2024splatam}:
\begin{equation}
\label{eq:rgb_loss}
    L_{pho} = \lambda_1 \left| C_k - \hat{C}_k \right| + \lambda_2 \left( 1 - \text{SSIM}(C_k, \hat{C}_k) \right),
\end{equation}
\begin{equation}
\label{eq:depth_loss}
    L_{geo}=|D_{k}-\hat{D}_{k}|,
\end{equation}
\begin{equation}
\label{eq:overall_loss}
    L=w_{c}L_{pho}+w_{d}L_{geo},
\end{equation}
where $\lambda_1=0.8, \lambda_2=0.2, w_{c} = 0.5, w_{d} = 1.0$ and $C_k, D_k$ are the captured RGB-D images of the $k$-th frame.

During the online mapping process, new Gaussians are dynamically initialized to cover newly observed areas, and redundant Gaussians with near-zero opacity or large covariances are removed as in~\cite{kerbl20233d}. Following~\cite{keetha2024splatam}, areas with low accumulated opacity or geometric deviations are identified as newly-observed areas:
\begin{equation}
\label{eq:add_gaussians}
    \begin{aligned}
    M_k& =\left(\hat{O}_k < \tau_{\mathrm{o1}} \right) \lor 
    \left((D_k<\hat{D}_{k}) \land (|D_{k}-\hat{D}_{k}|>\epsilon_{\mathrm{MDE}})\right),
    \end{aligned}
\end{equation}
where $\tau_{\mathrm{o1}} = 0.98$, and $\epsilon_{\mathrm{MDE}}$ represents $50$ times median depth error.

The dense prediction of the Gaussian map allows convenient extraction of the workspace and the obstacles. As illustrated in Box 3 of Fig.~\ref{fig: pipeline}, the top-down view can be efficiently rendered using a large focal length as the orthographic projection of the dense map. The region with sufficient accumulated opacity is taken as occupied, where areas above the ground and within the agent's height represent the obstacles. Navigable workspace can then be extracted as occupied areas at the ground level, excluding the obstacles. A Voronoi graph $\mathcal{G}=\{\mathcal{V},\mathcal{E}\}$ is generated through Voronoi tessellation~\cite{du1999centroidal}, with edges $\mathcal{E}$ being equidistant from the obstacles and nodes $\mathcal{N}$ as intersections where the edges terminate.

The Gaussian map and the Voronoi graph are complementary: The Gaussian map provides dense and complete information about previously visited regions of the scene, while the Voronoi graph offers a sparse structure of the workspace, which can also be seen as a strong deformation retract of the global free space~\cite{canny1987simplified} in topology. We show in the following that the integration of the two representations leads to an adaptive granularity of the environment and guarantees an effective trade-off between efficiency and accuracy during the autonomous reconstruction process.


\subsection{Active Viewpoint Selection}
\label{subsec:active_viewpoint_selection}
The objective of active mapping is to traverse the workspace and best capture the information of previously unseen areas. This is usually achieved by iteratively selecting target views, where the sampling strategy of accessible viewpoint candidates and the selection criteria are crucial to efficiency and overall coverage. We sample viewpoints on the Voronoi graph to maintain a compact and accessible set that covers the entire scene. Besides, the Voronoi graph generates a path that stays as far away from the obstacles as possible, thus guaranteeing a safe traversal~\cite{Kuang2024iros}.

\subsubsection{Decoupled Position and Rotation Candidates}
\label{subsubsec:position_rotation_candidates}
To best exploit the compact structure of the Voronoi graph along with the dense geometry and appearance of the Gaussian map, we propose decoupling the position and rotation candidates, rather than uniformly~\cite{xu2024hgs} or randomly~\cite{jin2025activegs} sampling viewpoints across the workspace. This approach reduces the search dimensionality while ensuring thorough observations. The dynamically updated map leads to a progressively refined graph that completely describes the partially-observed workspace. We iteratively select Voronoi nodes as viewpoint position candidates, where the node with the highest score (calculated as in Eq.~\ref{eq:node_score}) favors unvisited areas that push the boundary of the workspace for traversal. Regarding the view rotation, we adopt the yaw and pitch rotations at the selected view positions to get the observation toward a specific region. The target node position and the target rotation angle are determined in a view-dependent manner as follows.

\subsubsection{Coverage Evaluation}
\label{subsubsec:coverage_eval}
The actions regarding translation and rotation during the autonomous exploration process undergo different granularity. We aim to efficiently traverse the entire set of Voronoi nodes to maintain complete coverage, while conducting careful inspection in an area with intricate intersections of paths~\cite{Kuang2024iros}. In practice, we splat the constructed Gaussians to quickly obtain surrounding information of a specified pose, render panoramic images of the visibility at all unvisited nodes, as illustrated in Fig.~\ref{fig: pipeline}. To generate a sparse and representative set of rotation candidates, we apply the DBSCAN algorithm to cluster low-visibility regions (highlighted in red) within the panoramic view, where the pixel coordinate of the cluster center indicates yaw and pitch angles, as elaborated in Sec.~\ref{subsubsec:panorama_rendering}.

Note that the view-dependent accumulated opacity does not precisely reflect the reconstruction accuracy of the entire space. Firstly, the proportion of low-visibility areas in the image domain does not reflect the actual unseen space in three dimensions, as a node close to the unseen area results in a large amount of invisible pixels. Secondly, the accumulated opacity leads to an over-confident evaluation of completeness, as the splatting from backside geometry can also result in high accumulated opacity. To address these challenges, our method back-projects contour pixels, utilizing convex hull~\cite{barber1996quickhull} to get the approximated volume of the incomplete geometry. We further maintain a set of high-loss samples through joint analysis of observation depth $D_k$ and rendered depth $\hat{D}_k$:
\begin{equation}
\label{eq:high_loss_sample}
    \begin{aligned}
    M_h& = (\hat{O}_k > \tau_{\mathrm{o2}})  \land (D_k<\hat{D}_{k}) \land (|D_{k}-\hat{D}_{k}|>\epsilon_{\mathrm{1}}),
    \end{aligned}
\end{equation}
where $\tau_{\mathrm{o2}} = 0.8$, and $\epsilon_{\mathrm{1}} = 0.3$. The high-loss areas at each frame, before densification, are clustered with temporal propagation to keep track the newly observed regions.

\subsubsection{Determination of Target Views}
\label{subsubsec:view_selection}
The viewpoint selection is then conducted in two stages. The agent first selects the node with the most invisible areas, taking both panorama visibility measures and convex hull volumes into consideration. This strategy forces the agent to reach a closed space by fast marching the nodes with high scores, as detailed in Sec.~\ref{subsubsec:path_planning}, therefore expanding the workspace efficiently. Once the agent arrives at its goal position, the panorama image and the maintained high-loss samples guide the agent to rotate, as illustrated in Fig.~\ref{fig: pipeline} and Eq.~\ref{eq:high_loss_sample}, where invisible and high-loss areas get observations. The proposed method keeps the agent along the Voronoi graph that compresses potential viewpoints into a finite sparse set, reducing the computational complexity while guaranteeing completeness and safety.

\subsection{Hierarchical Planning with Voronoi Graph}
\label{subsec:hierarchical_plan}
\begin{figure}[t]
  \centering
  \includegraphics[width=0.85\linewidth]{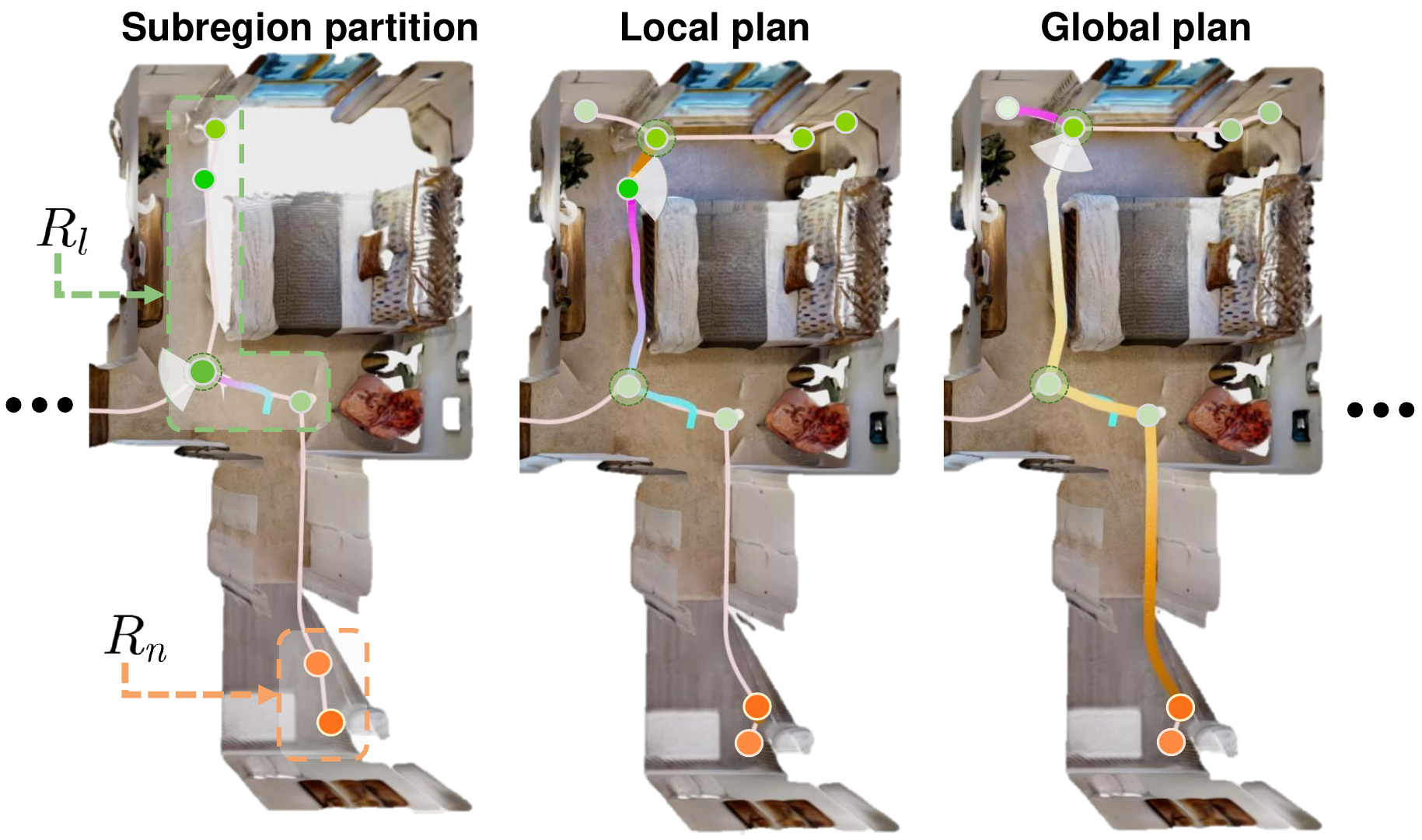}
  \caption{Once the agent gets sufficient observations within a local region $R_l$ (the green nodes), it selects the next sub-area $R_n$ within the highest score (the orange node) globally for further exploration.}
  \label{fig:subregion_partition}
\vspace{-0.2cm}
\end{figure}

In active mapping of multi-room environments, we observed that the prioritization of areas with high scores overlooks local coverage, resulting in paths that are unnecessarily visited multiple times, as shown in Fig.~\ref{fig:hierarchical_planning}. The nodes of a Voronoi graph represent distinct reachable regions, and the edges efficiently assess traversal costs between them. Therefore, we propose a hierarchical planning strategy based on Voronoi graphs, which includes subregion partitioning and local-global goal selection, as illustrated in Fig.~\ref{fig:subregion_partition}.

\subsubsection{Subregion Partition}
Building upon the topological structure of the Voronoi graph, we aim to dynamically partition it during the exploration into $n$ subregions $R_n$, including a local subregion $R_l$ where the agent is currently located, ensuring fine-grained local granularity with global guidance. In practice, we adopt the agglomerative hierarchical clustering method (UPGMA)~\cite{Arslan2016tro} for partitioning, with pairwise distances based on both Euclidean distance and travel path length. The hierarchy allows the flexibility to choose partitions at different levels and adapts to spatial data, without specifying the number of clusters beforehand.

\subsubsection{Local-Global Goal Selection}
The proposed method favors local exploration before global exploration. Once local areas are thoroughly explored, the next-best-subregion with the highest score is selected. Local planning is conducted by quantifying the incomplete score, as specified in Eq.~\ref{eq:node_score}, within the local subregion. The node with the highest score above a threshold is selected iteratively. Once the maximum score of the nodes within the local subregion falls below the threshold or all nodes are visited, the agent performs global planning by selecting the node outside the local subregion with the highest score. The global score not only takes the coverage into account, but also considers the distance cost along with the visited probabilities during past exploration. The active mapping process is then conducted to iterate between rigorous local mapping and coarse global exploration, to balance the reconstruction accuracy and efficiency.

\subsection{Implementation Details}
\label{subsec:details}
Given the selected target positions and rotations, the agent actively explores the unknown environment and captures new information. In the following, we discuss the details regarding bootstrapping, panorama rendering, path planning, and post-processing.
\subsubsection{Boostrapping}
Due to the limited field of view of the camera, we force the agent to look around at the very beginning. The agent takes discrete actions to execute 360 degrees of yaw rotation to obtain a complete ambient view.
\subsubsection{Panorama Rendering}
\label{subsubsec:panorama_rendering}
As the Gaussian splatting technique allows efficient rendering of pinhole images, we use three virtual cameras with $150$ degrees of FOV vertically and $120$ degrees of FOV horizontally to get the panoramic images, holistically quantifying the node-wise scores. The size of each panoramic image is set to be $360 \times 150$ to allow convenient selection of the rotation angle.
\subsubsection{Path Planning}
\label{subsubsec:path_planning}
Once the target goal position is determined, the shortest path can then be found through Dijkstra's algorithm. The score of the $i$-th node, denoted as $S_{i}$, is implemented as a weighted sum of the following factors:
\begin{equation}
\label{eq:node_score}
    \begin{aligned}
    S_{i} = w_o \cdot s_o(i) + w_c \cdot s_c(i) 
    + w_u \cdot s_u(i) + w_h \cdot s_h(i),
    \end{aligned}
\end{equation}
where $w_o = 20$, $w_c = 10$, $w_u = w_h = 10$. $s_o$ and $s_c$ are the portion of areas regarding the 2D invisible subregion and the 3D convex hull. $s_u$ and $s_h$ are the boolean values of unvisited and in-horizon states. Nodes with the same score are ranked according to their distance from the agent, where the nearer node is favored. We set a fixed distance threshold of 2 meters to control the granularity of subregion partitioning. We also enforce rotation once the agent arrives at multi-connected nodes in the graph as they are often intersection points between regions that require careful decision-making. Experiments indicate the efficacy of this strategy, achieving better efficiency for thorough exploration compared to previous state-of-the-art approaches\cite{Kuang2024iros}.

\subsubsection{Post-Processing}
\begin{figure}[t]
  \centering
  \includegraphics[width=0.85\linewidth]{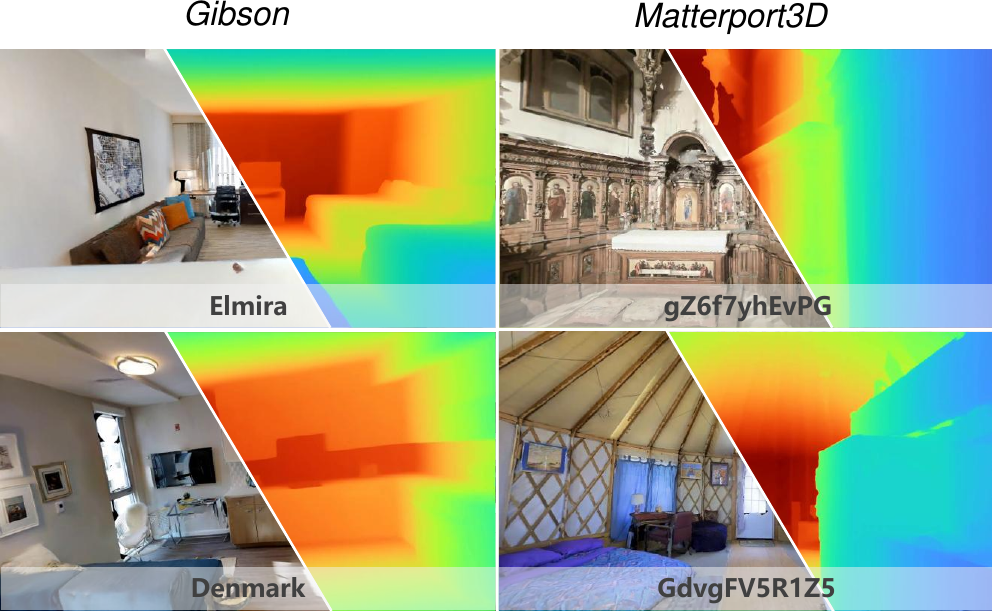}
  \caption{\textbf{Reconstruction results}: The autonomous reconstruction lead to photorealistic rendering and accurate geometry at a resolution of $512 \times 512$. The left and right sides of each image show rendered RGB and depth.}
  \label{fig:offline}
  \vspace{-0.5cm}
\end{figure}

Unlike NeRF-based SLAM algorithms that sacrifice model capacity for fast convergence to meet real-time demand, Gaussian splatting-based approaches maintain a consistent parameter space that allows post-processing. Therefore, we further apply adaptive density controls, as well as depth and normal regularization~\cite{kerbl20233d, Huang2DGS2024} to refine the online-constructed map given stored keyframe data. The reconstruction results are shown in Fig.~\ref{fig:offline}.
\section{Experiments}
\label{sec:experiments}
The experiments are conducted on a desktop PC with an Intel Core i9-12900K CPU and an NVIDIA RTX 3090 GPU. The reported results are averaged across 5 trials.
\subsection{Experimental Setup}
\label{subsec:experimental_setup}
To ensure fairness across exploration strategies, we perform qualitative and quantitative evaluations on the visually realistic Gibson~\cite{xiazamirhe2018gibsonenv} and Matterport3D~\cite{Matterport3D} datasets using the Habitat simulator~\cite{savva2019habitat}, following the protocol of~\cite{Yan2023iccv}. The single-floor scenes in the test/validation set are divided into small (less than 5 rooms) and medium (5–10 rooms) scenes. Unless otherwise specified, the agent collects posed RGB-D data at a resolution of $256 \times 256$ and performs discrete actions: \texttt{MOVE\_FORWARD} by 6.5 cm, \texttt{TURN\_LEFT} and \texttt{TURN\_RIGHT} by $10^{\circ}$, \texttt{TURN\_UP} and \texttt{TURN\_DOWN} by $15^{\circ}$, and \texttt{STOP}. The agent height is set to 1.25 m, with vertical and horizontal FOV of $90^{\circ}$. Additionally, the agent takes a $45^{\circ}$ downward pitch rotation to ensure a closed ground surface before departure.

\subsection{Evaluation Metrics}
\label{subsec:evaluation_metrics}
Following~\cite{feng2024naruto} and~\cite{Yan2023iccv}, we evaluate exploration coverage using completion ratio ($\%$) and completion ($cm$) across different map representations. To evaluate rendering quality, we use PSNR (dB), SSIM, and LPIPS for RGB rendering~\cite{kerbl20233d}, and Depth L1 ($cm$) distance for depth rendering performance. Additionally, we further evaluate the path length traveled by the agent during exploration.

\renewcommand{\arraystretch}{1.1}
\begin{table}[t]
    \caption{Comparison against relevant methods regarding the completeness of the observed data}
    \label{tab:method_compare}
    \centering
    \begin{tabular}{l@{\hspace{0.8cm}}c@{\hspace{0.8cm}}c@{\hspace{0.8cm}}c@{\hspace{0.8cm}}c}
        \toprule
        & \multicolumn{2}{c}{Gibson} & \multicolumn{2}{c}{Matterport3D} \\
        \cline{2-5}
        & \%↑ & $cm$↓ & \%↑ & $cm$↓ \\
        \toprule       \textbf{FBE}~\cite{yamauchi1997frontier} & 68.30 & 14.42 & 74.30 &9.29 \\
       \hline       \textbf{UPEN}~\cite{Georgakis2022icra} & 63.30 & 21.09 & 75.56 & 9.72\\
       \hline
       \textbf{ANM}~\cite{Yan2023iccv}  & 80.45 & 7.44 & 79.36 & 7.40 \\
       \hline
       \textbf{NARUTO}~\cite{feng2024naruto}  & 79.16 & 3.52 & 84.90 & 5.94 \\
       \hline
       \textbf{ANM-S}~\cite{Kuang2024iros} & 92.10 & 2.83 & 89.74 & 4.14 \\
       \hline
       \textbf{Ours} & \textbf{92.24} & \textbf{2.43} & \textbf{92.48} & \textbf{2.84}\\
       \bottomrule
    \end{tabular}
\vspace{-0.2cm}
\end{table}

\subsection{Comparison to Other Methods}
\label{subsec:comparison}
\begin{figure}[t]
  \centering
  \includegraphics[width=0.85\linewidth]{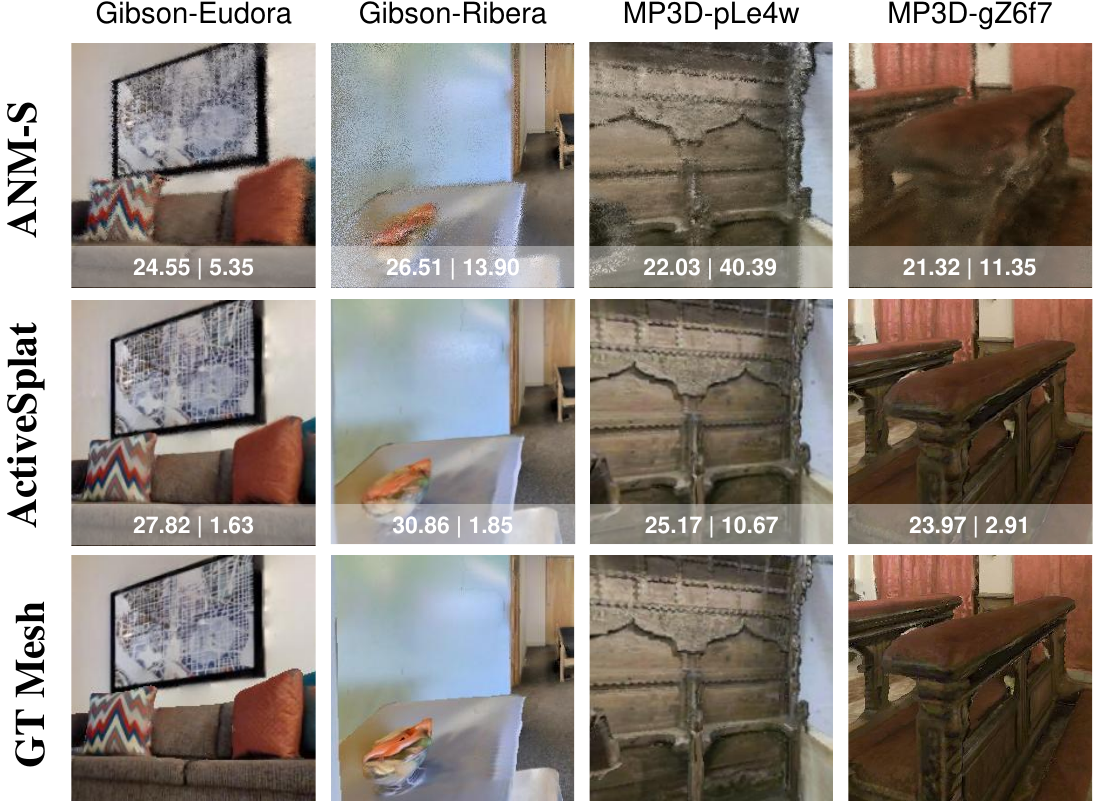}
  \caption{The novel view synthesis results of ours compared to the NeRF-based active mapping~\cite{Kuang2024iros} on Gibson and MP3D datasets. The bottom row of each picture shows the average PSNR (dB) and Depth L1 error ($cm$) of the scene from 50 randomly sampled test views.}
  \label{fig:quantitative_comparison}
\end{figure}

We first evaluate the exploration coverage across 13 different scenes following the setup of~\cite{Yan2023iccv}. As shown in Tab.~\ref{tab:method_compare}, the proposed system outperforms all relevant methods within the limited steps (1000 for small scenes and 2000 for medium-scale scenes). NARUTO~\cite{feng2024naruto} achieves $92\%$ coverage in simple scenes with a single room (e.g., MP3D-gZ6f7), thanks to its unrestricted movement capabilities, but is limited in complex scenes with multiple rooms (e.g., Cantwell and Eastville) due to its greedy strategy. Even though~\cite{Kuang2024iros} adopts a similar strategy of topology-guided exploration, the proposed method outperforms the baseline due to hierarchical planning that balances local reconstruction granularity and global scene coverage. We also performed a qualitative evaluation of novel view synthesis. As shown in Fig.~\ref{fig:quantitative_comparison}, the proposed method achieves better rendering quality of novel views when compared to the NeRF-based system~\cite{Kuang2024iros}, exhibiting sharp edges and ample textures, as detailed in the supplementary Sec.~\ref{subsec:rendering_comparison}.

\renewcommand{\arraystretch}{1.0}
\begin{table}[t]
    \caption{Ablation of exploration strategy}
    \label{tab:ablation_explore}
    \centering
    \begin{tabular}{l@{\hspace{1.2cm}}c@{\hspace{1.1cm}}c@{\hspace{1.1cm}}c@{\hspace{1.1cm}}c}
        \toprule
        & \multicolumn{2}{c}{Gibson} & \multicolumn{2}{c}{Matterport3D} \\
        \cline{2-5}
        & \%↑ & $cm$↓ & \%↑ & $cm$↓ \\
        \toprule
       \textbf{Random} & 84.20 & 6.13 & 83.91 &5.50 \\
       \hline
       \textbf{Position} & 90.41 & 2.74 & 89.54 & 3.67\\
       \hline
       \textbf{Viewpoint}  & 91.76 & \textbf{2.30} & 92.38 & 2.85 \\
       \hline
       \textbf{Ours} & \textbf{92.24} & 2.43 & \textbf{92.48} & \textbf{2.84}\\
       \bottomrule
    \end{tabular}
\vspace{-0.2cm}
\end{table}

\subsection{Ablation Study and System Performance}
\label{subsec:ablation_study}
To validate the rationale behind our solution, we conduct ablation studies and performance analyses to justify the effectiveness of different modules.

\subsubsection{Exploration Strategy}
\begin{figure*}[ht]
  \centering
  \includegraphics[width=0.8\linewidth]{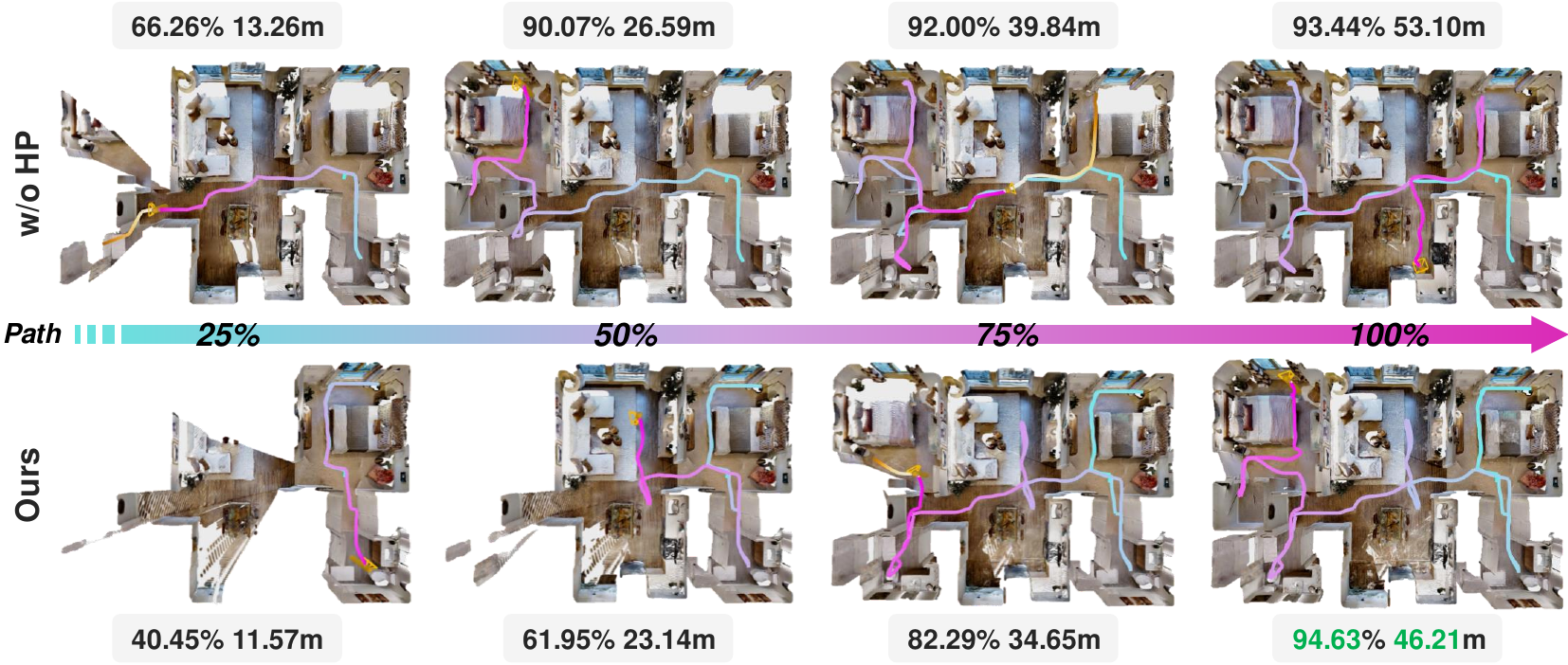}
  \caption{\textbf{Ablation of hierarchical planning (Gibson Quantico):} The online reconstruction progress with increased completion ratio ($\%$) and path length ($m$) at different stages. The hierarchical planning strategy results in better completeness and reduced path length during the exploration.}
  \label{fig:hierarchical_planning}
\end{figure*}

We first analyze the impact of different exploration strategies for thorough exploration, including a randomly sampled baseline (Random), exploration with node selection while ignoring target rotations (Position), decoupled selection of view positions and rotations (Viewpoint), and our proposed approach further incorporating multi-connected regions and hierarchical planning (Ours). As shown in Tab.~\ref{tab:ablation_explore}, different exploration strategies lead to diverse behaviors for efficiency-accuracy tradeoffs. The Random baseline indicates that the Voronoi graph guarantees complete exploration. Nevertheless, the traversal of all nodes without proper order is inefficient and overlooks certain areas. This issue remains for the greedy Position strategy as it only strives to push the boundaries of the workspace toward thorough traversal. The completeness can be improved with the rotation involved. We provide more detailed scene-wise results in the supplementary Sec.~\ref{subsec:scene_wise_robustness}.
Finally, the careful treatment of multi-connected nodes and the hierarchical planning (HP) strategy bring further advantages due to different inspection granularity locally and globally. Tab.~\ref{tab:ablation_hierarch_plan} shows the results of an ablation in which we increase the number of steps to 4000 (in all scenes) and compare the results with and without hierarchical planning at different stages. Our method benefits from fine-grained observations within local subregions, enabling higher completeness in the later stages. Additionally, we visualize the exploration trajectory when exploring a medium-sized environment. As illustrated in Fig.~\ref{fig:hierarchical_planning}, even though the greedy strategy leads to a rapid increase of completion in the beginning, coarsely exploring the neighboring areas results in repetitive trajectories.

\renewcommand{\arraystretch}{1.1}
\begin{table}[t]
    \caption{Ablation of hierarchical planning}
    \label{tab:ablation_hierarch_plan}
    \centering
    \resizebox{\linewidth}{!}{
    \begin{threeparttable}
    \begin{tabular}{lcccccccc}
        \toprule
        & \multicolumn{4}{c}{Gibson} & \multicolumn{4}{c}{Matterport3D} \\
        & \multicolumn{2}{c}{\textbf{w/o HP}} & \multicolumn{2}{c}{\textbf{Ours}}  & \multicolumn{2}{c}{\textbf{w/o HP}}  & \multicolumn{2}{c}{\textbf{Ours}} \\
        \cline{2-9}
        \textbf{Path Ratio} & $\%\uparrow$ & $cm\downarrow$ & $\%\uparrow$ & $cm\downarrow$ & $\%\uparrow$ & $cm\downarrow$ & $\%\uparrow$ & $cm\downarrow$ \\
        \toprule
       \textbf{25\%} & 87.48 & 4.96 & \textbf{88.73} & \textbf{4.49} & \textbf{92.61} & \textbf{2.73} & 91.50 & 3.04 \\ 
       \textbf{50\%} & 93.00 & 2.15 & \textbf{94.26} & \textbf{1.75} & \textbf{95.14} & \textbf{2.18} & 95.07 & 2.21 \\ 
       \textbf{75\%} & 95.75 & 1.20 & \textbf{96.20} & \textbf{1.04} & 95.28 & \textbf{2.14} & \textbf{95.31} & \textbf{2.14} \\ 
       \textbf{100\%} & 96.38 & 1.01 & \textbf{96.77} & \textbf{0.90} & 95.33 & \textbf{2.11} & \textbf{95.42} & \textbf{2.11} \\ 
       \bottomrule
    \end{tabular}
    \vspace{0.1cm}
    \begin{tablenotes}
        {\item[] \footnotesize
         Completion ratio (\%) and completion error ($cm$) under different allowed path lengths (expressed as percentages of the full trajectory).
        }
    \end{tablenotes}
    \end{threeparttable}
    }
\vspace{-0.2cm}
\end{table}

\renewcommand{\arraystretch}{1.0}
\begin{table}[t]
    \caption{Ablation of coverage evaluation}
    \centering
    \begin{tabular}{l@{\hspace{0.8cm}}c@{\hspace{0.8cm}}c@{\hspace{0.8cm}}c@{\hspace{0.8cm}}c}
        \toprule
        & \multicolumn{2}{c}{Gibson} & \multicolumn{2}{c}{Matterport3D} \\
        \cline{2-5}
        & \%↑ & $cm$↓ & \%↑ & $cm$↓ \\
        \toprule
       \textbf{Visibility only} & 90.19 & 3.09 & 91.40 &3.15 \\
       \hline
       \textbf{Convex hull only} & 91.09 & 2.86 & 91.50 & 2.98\\
       \hline
       \textbf{Ours} & \textbf{92.24} & \textbf{2.43} & \textbf{92.48} & \textbf{2.84}\\
       \bottomrule
    \end{tabular}
    \label{tab:ablation_visibility}
\vspace{-0.2cm}
\end{table}

\begin{table}[!t]
\caption{Ablation of post-processing on Gibson-Denmark}
\label{table:ablation_offline}
\centering
\resizebox{\linewidth}{!}{
\begin{tabular}{c|c|c|ccccc} 
\toprule
\multicolumn{1}{c|}{} & \textbf{Depth loss} &\textbf{Split}    & \textbf{Depth L1↓} & \textbf{PSNR↑}  & \textbf{SSIM↑} & \textbf{LPIPS↓} \\ 
\midrule
\multirow{2}{*}{\textbf{Online}} &  \multirow{2}{*}{True}                         & Train                & 1.91         & 25.28 & 0.83 & 0.22 \\ 
                       &                      & Test                 & 9.01 & 21.72 & 0.76 & 0.29 \\ 
\midrule
\multirow{4}{*}{\parbox{3em}{\textbf{Refined with 3DGS}}} & \multirow{2}{*}{False} & Train                & 4.49&\textbf{39.10}
&0.98&\textbf{0.03} \\ 
                       &                     & Test                 & 11.2&26.27&0.86&0.19 \\ 
\cline{2-7}
                       & \multirow{2}{*}{True}  & Train                & \textbf{0.77}&38.90&\textbf{0.99}&\textbf{0.03} \\ 
                       &                      & Test                 & 7.81 & 26.87 & \underline{0.88} & \underline{0.17} \\ 
\hline
\multirow{4}{*}{\parbox{3em}{\textbf{Refined with 2DGS}}} & \multirow{2}{*}{False} & Train                & 3.94 & 38.84 & \textbf{0.99} & 0.04 \\ 
                       &                      & Test                 & 10.10 & 27.13 & \underline{0.88} & 0.18 \\ 
\cline{2-7}
                       & \multirow{2}{*}{True}  & Train                & 0.80 & 38.90 & \textbf{0.99} & 0.04 \\ 
                       &                      & Test                 & \underline{7.56} & \underline{27.58} & \underline{0.88} & \underline{0.17}\\ 
\bottomrule
\end{tabular}
}
\vspace{-0.2cm}
\end{table}

\subsubsection{Coverage Evaluation}
As mentioned in Sec.~\ref{subsubsec:coverage_eval}, the quantification of visibility in the panoramic view guides the agent to push the boundary of the workspace. To verify the effectiveness of the integration of both the invisible mask area and convex hull volume, we compare different visibility quantification methods: guiding the agent to the node with the largest rendered area of invisible regions (Visibility only) and favoring the node with the largest 3D convex hull of the invisible boundary (Convex hull only). As shown in Tab.~\ref{tab:ablation_visibility}, using the view-dependent 2D results or the 3D volume quantification alone may not best evaluate the candidate nodes. The integration of both strategies (Ours) as weighted averaging of the normalized scores takes the relative extent of invisible areas near the Voronoi nodes into consideration, achieving promising completeness during the active mapping.

\subsubsection{Post-Processing}
The recent Gaussian splatting technique allows convenient post-processing with the stored keyframe buffer. We here compare our results before and after the post-processing using 3DGS~\cite{kerbl20233d} and 2DGS~\cite{Huang2DGS2024} on Gibson Denmark. $50$ frames of observations are selected uniformly as the train split, and $50$ images with randomly sampled camera poses within the free space are taken as the test split. As shown in Tab.~\ref{table:ablation_offline}, refinement can drastically enhance the reconstruction quality in terms of both geometry and appearance with the incorporation of RGB and depth during the optimization. The online autonomous exploration process allows active data capture for complete and high-fidelity reconstruction. It can be noted that the two-dimensional flattened Gaussian parameter representation of 2DGS~\cite{Huang2DGS2024}, along with the geometric regularization terms, shows better results in the test split, while 3DGS~\cite{kerbl20233d} indicates better overfitting in the train split. Besides, the use of depth loss, as defined in Eq.~\ref{eq:depth_loss}, not only leads to better geometry (lower Depth L1) in both train and test splits, but also enhances the generalization of the map (better quality in the test split). Refinement without depth loss may result in more realistic view synthesis results on training views (severe overfitting), but the geometry may deteriorate due to ambiguities in textureless areas.

\subsubsection{System Performance}
The average processing time per step for each module is presented in Tab.~\ref{tab:system_performance}. The proposed system runs at 8 fps in a headless mode, demonstrating real-time capable performance. Furthermore, the sparse decision-making ensures efficient viewpoint selection, while path planning constitutes only a small portion of the overall processing load. The major costs lie in the mapping and workspace extraction modules, which can be further accelerated with compressed Gaussian parameters.

\begin{table}[!t]
    \caption{Average Processing time per step}
    \centering
    \resizebox{\linewidth}{!}{
    \begin{tabular}{cccccccc}
        \toprule
        \textbf{Mapper} & \textbf{Get workspace} & \textbf{Get Voronoi} & \textbf{Get subregion} \\
        \hline
        45.14ms & 43.87ms & 0.69ms & 1.56ms \\ 
        \midrule
        \textbf{Rotation selection}  & \textbf{Position selection} & \textbf{Path planner} & \textbf{Visualizer (optional)} \\
        \hline
        2.76ms & 6.46ms & 0.20ms & 67.43ms \\
        \bottomrule 
    \end{tabular}
    }
    \label{tab:system_performance}
\end{table}

\begin{figure}[t]
  \centering
  \includegraphics[width=0.85\linewidth]{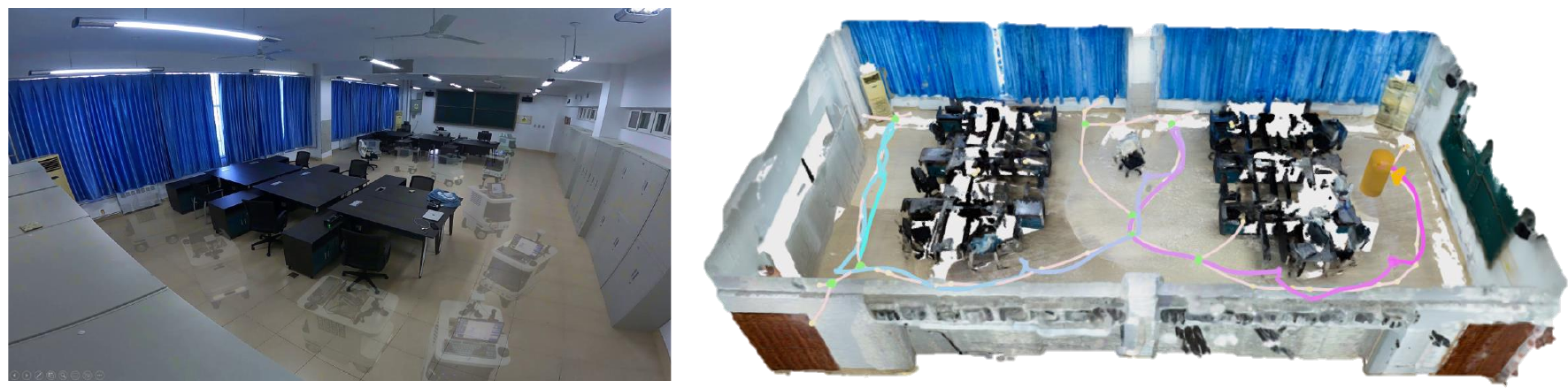}
  \caption{Real-world experimental result in an office scene.}
  \label{fig:real-world-experiment}
\vspace{-0.3cm}
\end{figure}

\subsection{Deployment in Real World}
To evaluate the practical usage of the proposed system in the real world, we deploy it on an omnidirectional mobile robot (Agile-X Ranger Mini) equipped with an Azure Kinect RGB-D sensor, where the camera pose is estimated by a line-based SLAM system~\cite{wang2021tro} in a parallel thread. Compared to simulation settings, we adjusted the radius hyperparameter of visited nodes for more detailed active mapping. The online reconstruction results are shown in Fig.~\ref{fig:real-world-experiment}, depicting the complete exploration that progressively and autonomously reconstructs the indoor environment. The detailed reconstruction process with additional results is provided in the supplementary video.

\section{Conclusion}
\label{sec:conclusion}
In this paper, we introduce an active mapping system for high-fidelity reconstruction of indoor scenes. Benefiting from the accurate dense prediction of a Gaussian splatting-based differentiable renderer and the workspace abstraction through Voronoi graph extraction, we employ the hybrid map along with a novel topology-based hierarchical planning strategy to achieve promising tradeoffs between exploration efficiency and completeness. Detailed experimental results indicate that the proposed system achieves effective tradeoffs between exploration coverage and reconstruction quality.

While ActiveSplat performs well, future work could further enhance its robustness, scalability, and real-world applicability. Integrating LiDAR into the Gaussian splatting-based SLAM system would enable robust tracking and more accurate geometric reconstruction in large-scale scenes. Moreover, high-fidelity exploration in unknown indoor environments opens new avenues for research in robotic autonomy, extending to tasks like lifelong navigation and mobile manipulation.

\addtolength{\textheight}{0.cm}   
\bibliographystyle{IEEEtran}
\bibliography{IEEEabrv,citations}
\vfill

\clearpage
\setcounter{page}{1} 
\maketitlesupplementary

\section{Further experimental analysis}

\subsection{Comparison of Rendering Quality with ANM-S}
\label{subsec:rendering_comparison}
We add the quantitative results regarding the rendering quality when compared against ANM-S~\cite{Kuang2024iros}. The evaluation follows the setup of Tab.~\ref{table:ablation_offline} at a rendered resolution of 512 × 512 for both systems, where the robot trajectories vary given different mapping strategies. Therefore, the training views are not the same, while the test views are taken from the same set through random sampling. Post-processing requires 30,000 iterations for both methods. The quantitative results are presented as shown in Tab.~\ref{tab:supp-quantitative_comparison}. We present the averaged quantitative results in Fig.~\ref{fig:quantitative_comparison} as a supplement. As mentioned in Sec.~\ref{sec:high_fidelity_scene_reconstruction}, Gaussian splatting-based representations allow more consistent optimization compared to NeRF-based representations, which sacrifice network capacity for efficiency, thereby achieving better rendering quality.

\renewcommand{\arraystretch}{1.0}
\begin{table*}[t]
    \caption{Comparison of rendering quality between our method and ANM-S on the Gibson and Matterport3D datasets, including Depth L1 [$cm$], PSNR [dB], SSIM, and LPIPS metrics.}
    \label{tab:supp-quantitative_comparison}
    \centering
    \resizebox{0.8\linewidth}{!}{
    \begin{tabular}{l|c|cccccccc}
        \toprule[1pt]
        \multirow{2}{*}{\textbf{Scene}} & \multirow{2}{*}{\textbf{Split}} & \multicolumn{2}{c}{\textbf{Depth L1↓}} & \multicolumn{2}{c}{\textbf{PSNR↑}}  & \multicolumn{2}{c}{\textbf{SSIM↑}}  & \multicolumn{2}{c}{\textbf{LPIPS↓}} \\
        \cline{3-10}
         & &ANM-S & Ours & ANM-S & Ours & ANM-S & Ours & ANM-S & Ours \\
        \midrule
        \multirow{2}*{\textbf{Gibson-Eudora}} & Train & 2.38 & \textbf{1.07} & 28.46 & \textbf{37.92} & 0.96 & \textbf{0.98} & 0.32 & \textbf{0.05} \\
        \cline{2-10}
        & Test & 5.34 & \textbf{1.63} & 24.55 & \textbf{27.82} & 0.89 & \textbf{0.90} & 0.44 & \textbf{0.16} \\
        \hline
        \multirow{2}*{\textbf{Gibson-Ribera}} & Train & 2.11 & \textbf{0.90} & 31.53 & \textbf{40.19} & 0.96 & \textbf{0.98} & 0.31 & \textbf{0.07} \\
        \cline{2-10}
        & Test & 13.90 & \textbf{1.85} & 26.51 & \textbf{30.86} & 0.88 & \textbf{0.91} & 0.42 & \textbf{0.17} \\
        \hline
        \multirow{2}*{\textbf{MP3D-pLe4w}} & Train & 6.27 & \textbf{1.54} & 27.60 & \textbf{33.81} & 0.91 & \textbf{0.94} & 0.32 & \textbf{0.13} \\
        \cline{2-10}
        & Test & 40.39 & \textbf{10.67} & 22.03 & \textbf{25.17} & 0.76 & \textbf{0.77} & 0.50 & \textbf{0.28} \\
        \hline
        \multirow{2}*{\textbf{MP3D-gZ6f7}} & Train & 5.31 & \textbf{0.93} & 24.62 & \textbf{33.82} & 0.91 & \textbf{0.96} & 0.31 & \textbf{0.09} \\
        \cline{2-10}
        & Test & 11.35 & \textbf{2.91} & 21.32 & \textbf{23.97} & \textbf{0.80} & 0.78 & 0.47 & \textbf{0.26} \\
        \bottomrule[1pt]
    \end{tabular}}
\end{table*}

\subsection{Scene-wise Ablation Study on Robustness}
\label{subsec:scene_wise_robustness}
Since the main paper only provides the mean values across all scenes, as shown in Tab.~\ref{tab:method_compare} and Tab.~\ref{tab:ablation_explore}, the overall variance is less informative due to the large performance gap among scenes. To better demonstrate the robustness of our method, we provide detailed results based on 5 ablation trials for each of the 13 scenes. The mean and standard deviation of the completion ratio for each scene are listed in Tab.\ref{table:supp-gibson_results} and Tab.~\ref{table:supp-mp3d_results}, respectively.

    \renewcommand{\arraystretch}{1.1}
    \begin{table*}[h]
    \caption{Experimental results on the Gibson dataset, including completion ratio (\%), completion ($cm$). Bold and underlined indicate the best and second-best results, respectively.}
    \label{table:supp-gibson_results}
    \centering
     \resizebox{1.0\linewidth}{!}{
    \begin{tabular}{c|c|ccccccccc|c}
    \toprule
    & \textbf{Metric} & \textbf{Denmark} & \textbf{Elmira} & \textbf{Eudora} & \textbf{Greigs.} & \textbf{Pablo} & \textbf{Ribera} & \textbf{Cantwell} & \textbf{Eastvi.} & \textbf{Swormv.} & \textbf{Avg.} \\
    \midrule
    \multirow{3}*{\textbf{Random}} & \%↑ & 92.72±0.43 & 90.77±1.90 & 89.52±3.34 & 89.42±2.26 & 71.67±1.71 & 85.82±1.36 & 77.31±5.13 & 75.74±4.50 & 78.16±4.70 & 83.46  \\
    & $cm$↓ & 2.01±0.19 & 2.64±0.45 & 2.67±0.81 & 3.39±0.75 & 13.99±3.02 & 4.48±0.67 & 9.31±2.84 & 9.34±2.95 & 7.52±2.89 & 6.15  \\
    \cline{2-12}
    
    \multirow{3}*{\textbf{Position}} 
    & \%↑ 
      & 93.91±0.67 & 92.25±\underline{0.98} & 92.80±0.25 & \textbf{98.96}±\textbf{0.02} & \underline{80.35}±\underline{1.07} & 86.85±\textbf{0.03} & \textbf{90.92}±0.64 & \underline{83.48}±5.90 & \underline{90.62}±1.36 & 90.02 \\
    & $cm$↓ 
      & 1.62±0.16 & 2.31±\underline{0.16} & 1.74±\underline{0.04} & \textbf{0.60}±\textbf{0.00} & \underline{5.55}±0.78 & 5.32±\underline{0.03} & \textbf{2.24}±\underline{0.11} & \underline{5.93}±2.68 & \underline{1.97}±0.31 & 3.03 \\
    \cline{2-12}
    
    \multirow{3}*{\textbf{Viewpoint}} & \%↑ & \textbf{95.19}±\textbf{0.04} & \underline{94.07}±2.21 & \textbf{94.17}±\underline{0.05} & \underline{98.33}±\textbf{0.02} & 79.89±\textbf{0.08} & \underline{94.62}±0.30 & 86.33±\textbf{0.11} & 83.33±\textbf{3.25} & 90.53±\underline{0.95} & \underline{90.72}  \\
    & $cm$↓ & \textbf{1.35}±\textbf{0.01} & \underline{1.85}±0.67 & \textbf{1.39}±0.05 & \underline{0.74}±\textbf{0.00} & 6.05±\textbf{0.01} & \underline{1.26}±0.06 & 4.80±\textbf{0.03} & \underline{5.93}±\textbf{1.61} & 2.07±\underline{0.27} & \underline{2.83}  \\
    \cline{2-12}
    
    \multirow{3}*{\textbf{Ours}} & \%↑ & \underline{94.63}±\underline{0.28} & \textbf{97.70}±\textbf{0.16} & \underline{93.83}±\textbf{0.03} & 94.55±\underline{0.15} & \textbf{85.83}±1.91 & \textbf{96.42}±\underline{0.10} & \underline{86.99}±\underline{0.56} & \textbf{87.12}±\underline{4.09} & \textbf{92.04}±\textbf{0.84} & \textbf{92.12}\\
    & $cm$↓ & \underline{1.57}±\underline{0.11} & \textbf{0.86}±\textbf{0.03} & \underline{1.50}±\textbf{0.01} & 1.60±\underline{0.02} & \textbf{3.47}±\underline{0.45} & \textbf{0.96}±\textbf{0.02} & \underline{4.03}±0.98 & \textbf{4.25}±\underline{1.91} & \textbf{1.90}±\textbf{0.23} & \textbf{2.24} \\
    \bottomrule
    \end{tabular}
    }
    \end{table*}

    \renewcommand{\arraystretch}{1.1}
    \begin{table*}[h]
    \caption{Experimental results on the MP3D dataset, including completion ratio (\%), completion ($cm$). Bold and underlined indicate the best and second-best results, respectively.}
    \label{table:supp-mp3d_results}
    \centering
    \resizebox{0.7\linewidth}{!}{
    \begin{tabular}{c|c|cccc|c}
    \toprule
    & \textbf{Metric} & \textbf{gZ6f7yhEvPG} & \textbf{pLe4wQe7qrG} & \textbf{GdvgFV5R1Z5} & \textbf{YmJkqBEsHnH} & \textbf{Avg.} \\
    \midrule
    \multirow{3}*{\textbf{Random}} & \%↑ & 90.29±0.63 & 85.78±4.95 & 85.76±3.77 & 79.97±3.66 & 85.45 \\
    & $cm$↓ & 3.27±0.46 & 5.09±1.66 & 5.15±1.05 & 6.20±1.69 & 4.93 \\
    \cline{2-7}
    \multirow{3}*{\textbf{Position}} & \%↑ & 90.99±\underline{0.58} & 93.96±0.75 & 90.56±0.10 & \textbf{86.68}±\underline{0.50} & 90.55 \\
    & $cm$↓ & 3.44±0.28 & 2.23±0.17 & \underline{3.98}±\underline{0.03} & \textbf{3.70}±\underline{0.18} & 3.34 \\
    \cline{2-7}
    \multirow{3}*{\textbf{Viewpoint}} & \%↑ & \textbf{96.35}±0.71 & \underline{96.95}±\textbf{0.24} & \textbf{91.85}±\underline{0.08} & 82.00±1.37 & \underline{91.79} \\
    & $cm$↓ & \textbf{1.49}±\underline{0.06} & \underline{1.47}±\textbf{0.04} & \textbf{3.81}±\textbf{0.02} & 4.88±0.39 & \underline{2.92} \\
    \cline{2-7}
    \multirow{3}*{\textbf{Ours}} & \%↑ & \underline{95.91}±\textbf{0.30} & \textbf{97.09}±\underline{0.41} & \underline{91.78}±\textbf{0.06} & \underline{83.91}±\textbf{0.48} & \textbf{92.17} \\
    & $cm$↓ & \underline{1.61}±\textbf{0.04} & \textbf{1.44}±\underline{0.07} & \textbf{3.81}±\textbf{0.02} & \underline{4.39}±\textbf{0.16} & \textbf{2.82} \\
    \bottomrule
    \end{tabular}
    }
    \end{table*}

\subsection{Supplementary Analysis on Coverage Evaluation}
\label{subsec:supplementary_analysis}
\begin{figure*}[t]
  \centering
  \includegraphics[width=0.7\linewidth]{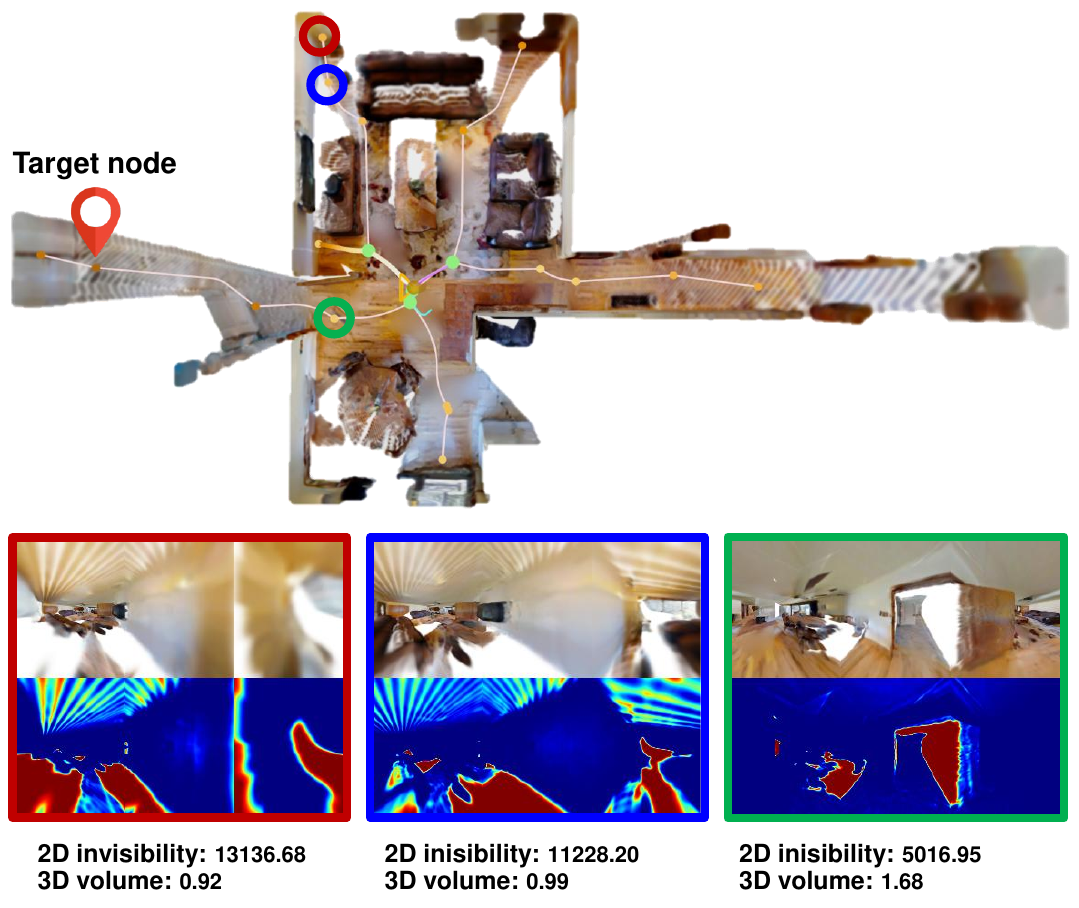}
  \caption{The rendered panoptic images and the opacity maps for the node scoring.}
  \label{fig:supp-2d-3d-cantwell}
\end{figure*}

Here, in supplementary to Tab.~\ref{tab:ablation_visibility}, we provide one empirical result to validate the design. As shown in Fig.~\ref{fig:supp-2d-3d-cantwell}, the nodes in red and blue have higher invisibility scores compared to the green node, as they lie closer to the invisible areas. On the contrary, the green node has a higher 3D volume score of the corresponding convex hull compared to the red and blue nodes, as it lies near a larger area of open space to be explored. This indicates the biased scoring through view-based rendering criteria. Nonetheless, for the red and blue nodes that share similar non-visited areas, we favor the node that is closer to the areas, which can be reflected as a higher 2D invisibility score. Despite this empirical design, the selected target node usually has significantly higher scores compared to other nodes. Either case may not significantly affect the final results.

\begin{figure*}[t]
  \centering
  \includegraphics[width=0.75\linewidth]{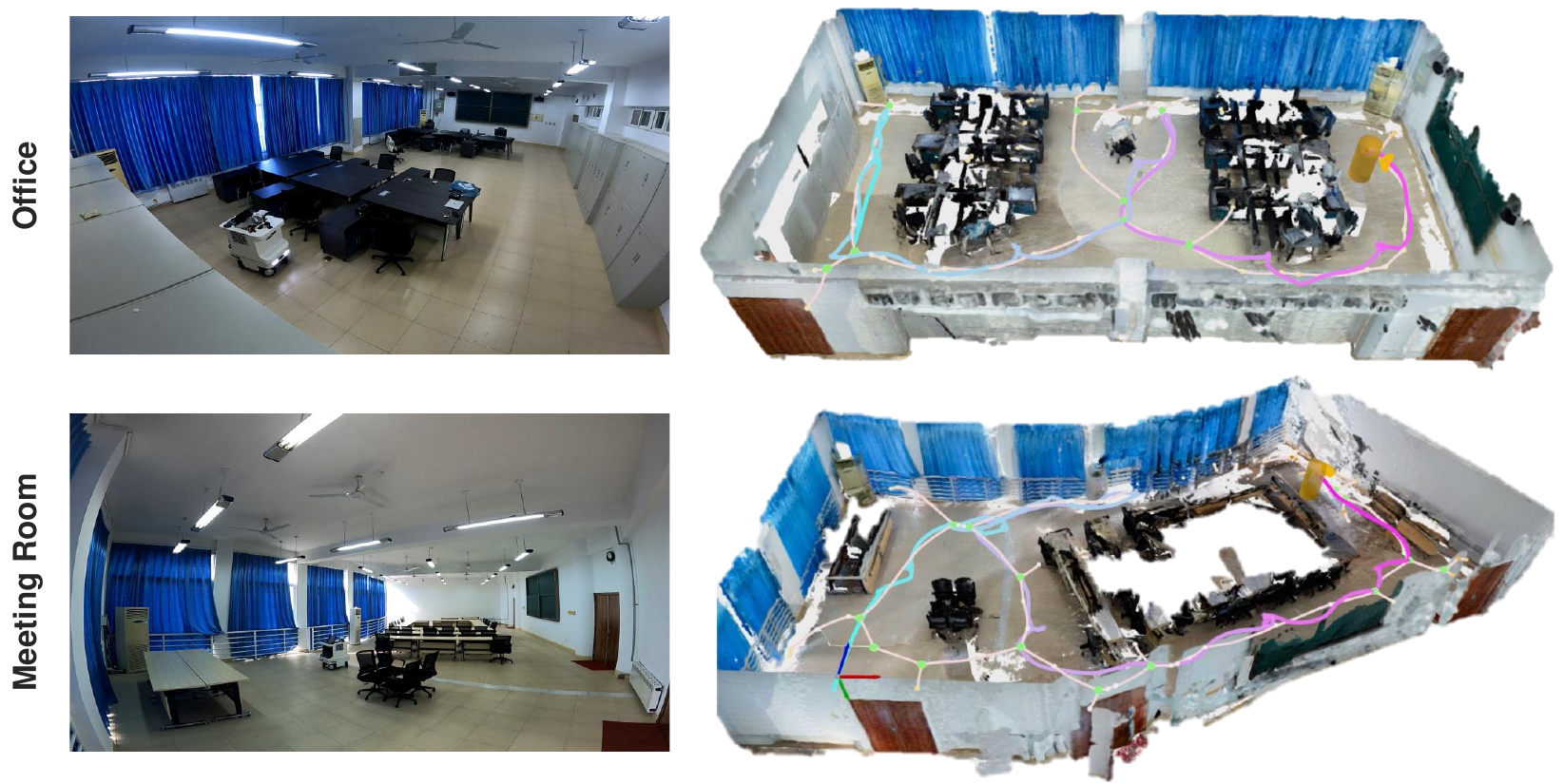}
  \caption{Real-world experimental results in an office and a meeting room.}
  \label{fig:supp-real-world-experiment}
\end{figure*}

\end{document}